\documentclass[sigconf,authorversion,nonacm]{acmart}

\AtBeginDocument{%
  \providecommand\BibTeX{{%
    \normalfont B\kern-0.5em{\scshape i\kern-0.25em b}\kern-0.8em\TeX}}}

\usepackage{xspace}
\usepackage{enumerate}
\usepackage{amsmath}
\usepackage[capitalise,noabbrev]{cleveref}

\usepackage[inline]{enumitem}

\usepackage{tikz}
\usepackage{tabularx,environ}
\pgfdeclarelayer{background}
\pgfsetlayers{background}
\usepackage{float}
\usepackage{subcaption} 
\usepackage{multirow}
\usepackage{nicefrac}
\usepackage{combelow}
\usepackage{spreadtab}
\usepackage{booktabs}
\usepackage{enumitem}
\usepackage{colortbl}
\usepackage{footnote}
\makesavenoteenv{tabular}
\usepackage[utf8]{inputenc}
\usepackage{textcomp}
\usepackage{caption}
\usepackage{pgfplots}
\usepackage{adjustbox}
\usepackage{amsthm}

\usetikzlibrary{shapes.geometric, arrows}
\usetikzlibrary{backgrounds}

\newif\ifdraft\drafttrue
\newif\ifreviewChanges\reviewChangestrue
\newif\ifinlineref\inlinereffalse
\newif\iffinal\finalfalse
\newif\ifextended\extendedfalse
\newif\ifdotikz\dotikzfalse
\newif\ifmakeallproofsinline\makeallproofsinlinefalse
\definecolor{blue(pigment)}{rgb}{0.2, 0.2, 0.6}

\draftfalse
\finaltrue

\ifdraft

\newcommand{\me}[1]{{\textbf{\textcolor{purple}{ [M: #1 ] }}}}
\newcommand{\kien}[1]{\textcolor{orange}{[K: #1]}} 
\newcommand{\ds}[1]{\textcolor{blue}{[DS: #1]}}

\newcommand{\add}[1]{\textcolor{blue(pigment)}{#1}}
\definecolor{darkgreen}{rgb}{0.0, 0.5, 0.0}

\else

\newcommand{\me}[1]{}
\newcommand{\kien}[1]{} 
\newcommand{\ds}[1]{}
\newcommand{\add}[1]{#1}

\fi

\ifreviewChanges
\newcommand{\dl}[1]{{\color{darkred}\sout{#1}}}
\else
\newcommand{\dl}[1]{}
\fi

\newcommand{\dlliteR}{\ensuremath{\mathit{DL}\text{-}\mathit{Lite}_{\mathcal{R}}}\xspace}

\newcommand{\mi}[1]{\mathit{#1}}
\newcommand{\cE}{\ensuremath{\mathcal{E}}}

\newcommand{\cI}{\ensuremath{\mathcal{I}}}
\newcommand{\bC}{\ensuremath{\mathbf{C}}}
\newcommand{\bI}{\ensuremath{\mathbf{E}}}
\newcommand{\bR}{\ensuremath{\mathbf{R}}}
\newcommand{\cO}{\ensuremath{\mathcal{O}}}

\newcommand{\cR}{\ensuremath{\mathcal{R}}}
\newcommand{\Omc}{\ensuremath{\mathcal{O}}}

\newcommand{\Gmc}{\ensuremath{\mathcal{G}}}
\newcommand{\Imc}{\ensuremath{\mathcal{I}}}

\newcommand{\Qmc}{\ensuremath{\mathcal{Q}}}

\newcommand{\tuple}[1]{\ensuremath{\langle #1 \rangle}}

\newcommand{\Vbf}{\ensuremath{\mathbf{V}}}

\newcommand{\ISA}{\ensuremath{\sqsubseteq}}

\newcommand{\vars}{\ensuremath{\mathit{vars}}}

\newcommand{\cbf}{\ensuremath{\mathbf{c}}}
\newcommand{\qbf}{\ensuremath{\mathbf{q}}}

\newcommand{\cen}{\ensuremath{\mathbf{cen}}}
\newcommand{\off}{\ensuremath{\mathbf{off}}}

\newcommand{\rel}[1]{\ensuremath{\mathrm{#1}}}
\newcommand{\ent}[1]{\ensuremath{\mathrm{#1}}}

\newcommand{\plain}{\ensuremath{\mathit{Plain}}}
\newcommand{\spe}{\ensuremath{\mathit{Spe}}}
\newcommand{\gen}{\ensuremath{\mathit{Gen}}}
\newcommand{\onto}{\ensuremath{\mathit{Onto}}}

\newcommand{\tstA}{{\bf I}}
\newcommand{\tstB}{{\bf D}}
\newcommand{\tstC}{{\bf I+D}}

\tikzstyle{rectangleNode} = [rectangle, rounded corners, minimum width=1cm, minimum height=1cm,text centered, draw=black, thick]
\tikzstyle{diamondNode} = [diamond, rounded corners, minimum width=0.5cm, minimum height=0.5cm, text centered, draw=black, thick]
\tikzstyle{arrow} = [thick,->,>=stealth]

\makeatletter
\newcommand{\problemtitle}[1]{\gdef\@problemtitle{#1}}
\newcommand{\probleminput}[1]{\gdef\@probleminput{#1}}
\newcommand{\problemquestion}[1]{\gdef\@problemquestion{#1}}
\NewEnviron{problemQ}{
	\problemtitle{}\probleminput{}\problemquestion{}
	\BODY
	\par\addvspace{.5\baselineskip}
	\noindent
	\begin{tabularx}{\textwidth}{@{\hspace{\parindent}} l X c}
		\multicolumn{2}{@{\hspace{\parindent}}l}{\@problemtitle} \\
		\textbf{Input:} & \@probleminput \\
		\textbf{Question:} & \@problemquestion
	\end{tabularx}
	\par\addvspace{.5\baselineskip}
}
\NewEnviron{problemO}{
	\problemtitle{}\probleminput{}\problemquestion{}
	\BODY
	\par\addvspace{.5\baselineskip}
	\noindent
	\begin{tabularx}{\textwidth}{@{\hspace{\parindent}} l X c}
		\multicolumn{2}{@{\hspace{\parindent}}l}{\@problemtitle} \\
		\textbf{Input:} & \@probleminput \\
		\textbf{Output:} & \@problemquestion
	\end{tabularx}
	\par\addvspace{.5\baselineskip}
}

\makeatother

\newcommand{\qbplain}{\ensuremath{\mathit{Q2B}_\mathit{plain}}\xspace}

\newcommand{\qbox}{\ensuremath{\mi{Q2B}}\xspace}
\newcommand{\qboxrew}{\ensuremath{\mi{Q2B}_{\mi{plain}}^{\mi{rew}}}\xspace}

\newcommand{\cqd}{\ensuremath{\mi{CQD}}\xspace}
\newcommand{\cqdasr}{\ensuremath{\mi{CQD}^{\mi{ASR}}}\xspace}

\newcommand{\cqdplain}{\ensuremath{\mi{CQD}_{\mi{plain}}\xspace}}
\newcommand{\cqdrew}{\ensuremath{\mi{CQD}_{\mi{plain}}^{\mi{rew}}}\xspace}
\newcommand{\obox}{\ensuremath{\mi{O2B}}\xspace}

\usepackage{xspace}
\newcommand*{\eg}{e.g.\@\xspace}
\newcommand*{\ie}{i.e.\@\xspace}

\makeatletter
\newcommand*{\etc}{%
    \@ifnextchar{.}%
        {etc}%
        {etc.\@\xspace}%
}
\makeatother

\usepackage{amsmath,amsfonts,bm}

\def\eqref#1{equation~\ref{#1}}

\def\1{\bm{1}}

\DeclareMathAlphabet{\mathsfit}{\encodingdefault}{\sfdefault}{m}{sl}
\SetMathAlphabet{\mathsfit}{bold}{\encodingdefault}{\sfdefault}{bx}{n}

\makeatletter
\def\@copyrightspace{\relax}
\makeatother

\begin{document}


\title{Combining Inductive and Deductive Reasoning for Query Answering over Incomplete Knowledge Graphs}

\author{Medina Andresel}\authornote{The work has been done during the PhD sabbatical of Medina Andresel at Bosch Center for Artificial Intelligence}
\affiliation{%
\institution{AIT Austrian Institute of Technology}
  \country{Vienna, Austria}}
\email{medina.andresel@ait.ac.at}

\author{Trung-Kien Tran}
\affiliation{%
  \institution{Bosch Center for Artificial Intelligence}
  \country{Renningen, Germany}
}
\email{trungkien.tran@de.bosch.com}

\author{Csaba Domokos}
\affiliation{%
  \institution{Bosch Center for Artificial Intelligence}
  \country{Renningen, Germany}}
\email{csaba.domokos@de.bosch.com}

\author{Pasquale Minervini}
\affiliation{%
  \institution{University of Edinburgh}
  \country{Edinburgh, United Kingdom}
}
\email{p.minervini@ed.ac.uk}

\author{Daria Stepanova}
\affiliation{%
 \institution{Bosch Center for Artificial Intelligence}
 \country{Renningen, Germany}
 }
  \email{daria.stepanova@de.bosch.com}

\renewcommand{\shortauthors}{Medina Andresel, Trung-Kien Tran, Csaba Domokos, Pasquale Minervini, \& Daria Stepanova}

\begin{abstract}
Current methods for embedding-based query answering over incomplete Knowledge Graphs (KGs) only focus on inductive reasoning, i.e., predicting answers by learning patterns from the data, and lack the complementary ability to do deductive reasoning,
which requires the application of domain knowledge to infer further information.
To address this shortcoming, we investigate the problem of incorporating ontologies into embedding-based query answering models by defining the task of embedding-based ontology-mediated query answering. We propose various integration strategies into prominent representatives of embedding models that involve (1) different ontology-driven data augmentation techniques and (2) adaptation of the loss function to enforce the ontology axioms. We design novel benchmarks for the considered task based on the LUBM and the NELL KGs and evaluate our methods on them. The achieved improvements in the setting that requires both inductive and deductive reasoning are from 20\% to 55\% in HITS@3.
\end{abstract}



\maketitle

\section{Introduction}
\begin{figure}[t]
\begin{center}
\includegraphics[scale=.24]{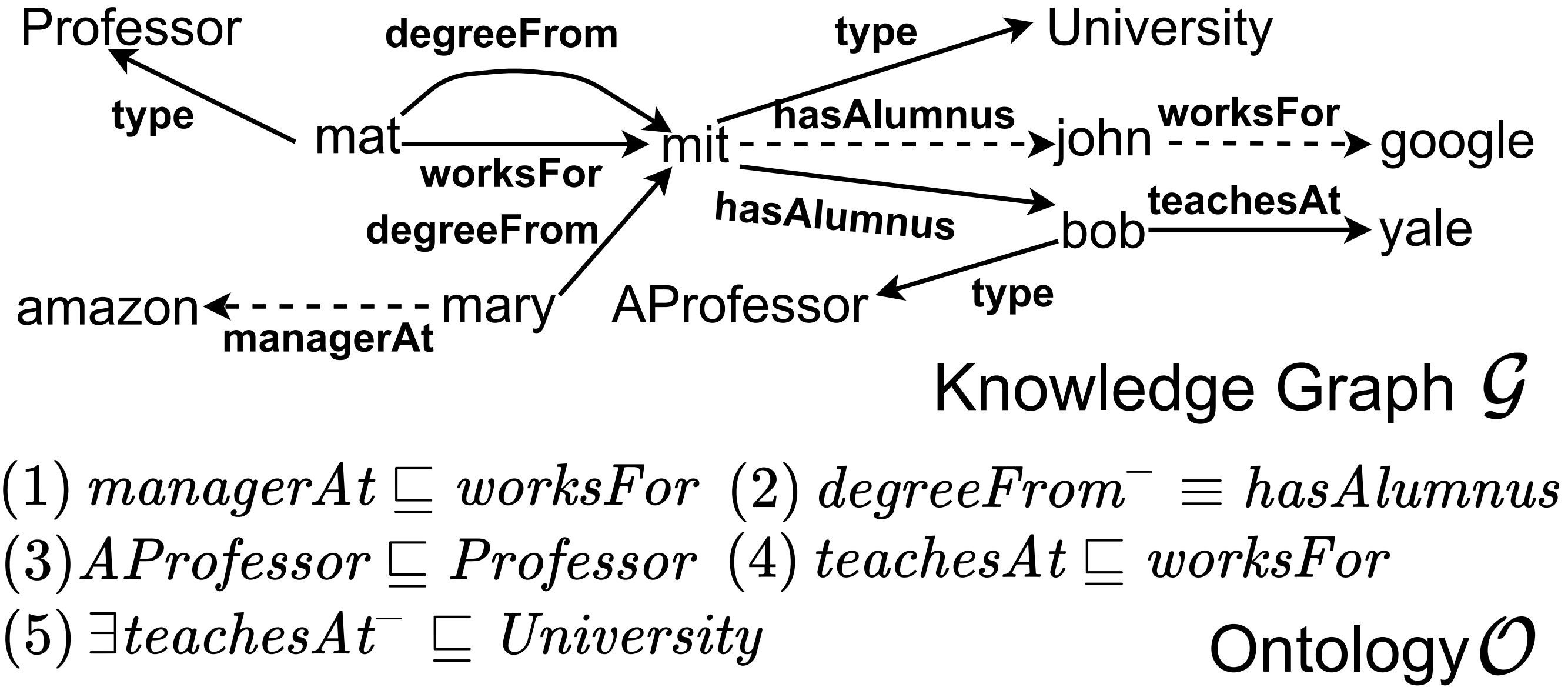}
\caption{An exemplary ontology $\Omc$ and a KG $\Gmc$. Solid edges illustrate existing facts in $\Gmc$, and dashed ones indicate missing facts that could be predicted using KG embeddings. 
}\label{fig:kgonto} 
\end{center}
\end{figure}

Knowledge Graphs (KGs) have recently received much attention due to their relevance in various applications, such as natural question answering or web search. Prominent KGs include NELL~\cite{nell}, YAGO~\cite{yago}, and Wikidata~\cite{wikidata}. A KG describes facts about entities by interconnecting them via  relations, \eg, $\mi{hasAlumnus(mit, bob)}$ in  \cref{fig:kgonto} 
states that Bob is an MIT alumnus. 

A crucial task in leveraging 
information from knowledge graphs is that of answering logical queries 
such as \textit{Who works for Amazon and has a degree from MIT?}, which can be formally written as  \\$q(X)\leftarrow  \mi{degreeFrom}(X, \mi{mit}) \wedge \mi{worksFor}(X, \mi{amazon})$.
Answering such queries is very challenging when 
KGs are incomplete, which is often the case due to their (semi-) automatic construction, and obtaining 
complete answers typically requires further 
domain knowledge, \ie, the application of deductive reasoning. For instance, $\mi{mary}$ is a missing but desired answer of $q$ that can be obtained by combining \add{the fact $\mi{managerAt(mary,amazon)}$, predicted 
using machine learning models, and the axiom stating that $\mi{managerAt}$ implies $\mi{worksFor}$ in the ontology $\Omc$ of \cref{fig:kgonto}.} Therefore, such a task requires both inductive and deductive reasoning.

Recently, \emph{Knowledge Graph Embedding} (KGE) techniques~ \cite{DBLP:journals/pieee/Nickel0TG16} that can be used to predict missing facts have been proposed to answer logical queries over incomplete KGs~\cite{DBLP:conf/nips/HamiltonBZJL18,iclr/RenHL20,DBLP:conf/nips/RenL20,DBLP:conf/nips/SunAB0C20,DBLP:conf/kdd/LiuDJZT21}. At the same time, in the Knowledge Representation and Reasoning area answering queries over incomplete data has also received a lot of attention and one of the most successful approaches for this task is to exploit ontologies when querying KGs, referred to as \emph{Ontology-Mediated Query Answering} \cite[OMQA,][]{DBLP:journals/ki/SchneiderS20a}.

While promising, existing embedding-based methods do not take ontologies, which formalize domain knowledge, into account. Since large portions of expert knowledge can be conveniently encoded using ontologies, the benefits of coupling ontology reasoning and embedding 
methods for KG completion are evident and have been acknowledged in several works,  \eg~\cite{kr/Gutierrez-Basulto18,DBLP:journals/corr/abs-1902-10499}. 
However, to the best of our knowledge, coupling inductive and deductive reasoning to answer queries over incomplete KGs
has not been considered yet.

Answering queries over the
KG augmented 
with triples resulting from the naive process of interchangeably using embedding methods and ontology reasoning, comes with a big scalability challenge~\cite{KrompassNT14} and commonly known error accumulation issues. In practice, we need to restrict ourselves to computing merely small subsets of likely fact predictions required for answering a given query; 
thus more sophisticated proposals are needed. 
Hence, we investigate three open questions: (1) How to adapt existing OMQA techniques to the setting of KGEs?
(2) How do different data augmentation strategies impact the accuracy of  existing embedding models for the OMQA task? 
(3) Does enforcing ontology axioms in the embedding space via loss function help to improve inductive and deductive reasoning performance?
\newpage
We answer (1)-(3) 
by the following contributions:
\begin{itemize}[leftmargin=*] \itemsep1pt
    \item We formally define the novel task of \emph{Embedding-Based Ontology-Mediated Query Answering} (E-OMQA), analyze and systematically compare various extensions of embedding-based query answering models to incorporate ontologies. 
    \item We propose ontology-driven strategies for sampling queries to train embedding models for query answering, as well as a loss function modification to enforce the ontology axioms within the embedding space, and demonstrate the effectiveness of these proposals on widely-used representatives of query-based and atom-based models. 
    \item As no previous benchmarks exist for E-OMQA, we design new ones using LUBM and NELL, i.e., well-known benchmarks for OMQA and embedding models, respectively. 
    \item Extensive evaluation shows that enforcing the ontology via the loss function, in general, improves the deductive power regardless of how the training data is sampled, while ontology-driven sampling strategy has a further significant positive impact on performance.
    We obtain overall improvements, ranging from 20\% to 55\% in HITS@3, in the settings that require both inductive and deductive reasoning.  
\end{itemize}

\section{Preliminaries}\label{sec:prelims}
\paragraph{Knowledge Graphs and Ontologies}
We assume a signature consisting of countable pairwise disjoint sets $\bI,\bC$, and $\bR$ of entities (constants), concepts (types), and roles (binary relations), respectively. A knowledge graph $\Gmc$ (\emph{a.k.a.} ABox) is a  set of triples, such as $\mi{(mit, type, University)}$ and $\mi{(bob, worksFor, mit)}$, where $\mi{mit},\mi{bob} \in \bI$, $\mi{worksFor}, \mi{type} \in \bR$, and $\mi{University} \in \bC$. These triples can also be represented as $\mi{University(mit)}$\footnote{Unary facts 
can also be modeled using the binary $\mi{type}$ relation.} and $\mi{worksFor(bob, mit)}$.
An ontology $\cO$ (\emph{a.k.a.} TBox) is a set of axioms 
in Description Logics~\cite{DBLP:series/ihis/BaaderHS09} over the signature $\Sigma = \tuple{\bI, \bC,\bR}$. 
We focus on ontologies in $\mi{DL}$-$\mi{Lite}_{\cR}$ DL fragment~\cite{DBLP:journals/jar/CalvaneseGLLR07} that have the following syntax:

\small{
\begin{table}[t]
\caption{DL syntax and semantics defined using FO interpretations $(\Delta^{\cI}, \cdot^\cI)$ with a non-empty domain $\Delta^{\cI}$ and an interpretation function $\cdot^{\cI}$. $C$ and $D$ denote concepts in $\mi{DL}$-$\mi{Lite}_{\cR}$.}
\label{tab:dl}
\begin{tabular}{ll}
\toprule
\textbf{DL Syntax}     & \textbf{Semantics}                             \\
\midrule
$e$ & $e^{\cI} \in \Delta^{\cI}$\\
$A$ (resp. $p$) & $A^{\cI} \subseteq \Delta^{\cI}$ (resp. $p^{\cI}\subseteq \Delta^{\cI} {\times} \Delta^{\cI}$)\\
$\exists p$                                 & $(\exists p)^{\Imc} = \{d\in\Delta^{\cI}\,|\,\exists d', (d,d')\in p^{\cI}\}$ \\
$p^-$          & $(p^-)^\Imc=\{(d',d)\mid (d,d') \in p^\Imc\}$. \\
$C \sqsubseteq D$ (resp. $p \sqsubseteq s$) & $C^{\cI}\subseteq D^{\cI}$ (resp. $p^{\cI} \subseteq s^{\cI}$)                \\
$A(c)$ (resp. $p(c,c')$) & $c \in A^\cI$ (resp. $(c,c') \in p^\cI$)     \\  
\bottomrule
\end{tabular}
\end{table}}
\normalsize 

$\begin{aligned}
A &\ISA A' &A &\ISA \exists p &\exists p &\ISA A 
&\exists p^- &\ISA A &p &\ISA  s &p^- &\sqsubseteq s,
\end{aligned}$

\noindent where $A,$ $A'\in \bC$ are concepts and $p,s\in \bR$ are roles and $p^-$ denotes the inverse relation of $p$.  
The KG and its ontology from Figure~\ref{fig:kgonto} are in $\mi{DL}$-$\mi{Lite}_{\cR}$. The $\mi{DL}$-$\mi{Lite}_{\cR}$ syntax and semantics are summarized in Table~\ref{tab:dl}. Given a KG $\Gmc$ and an ontology $\Omc$, an interpretation $\cI$ \emph{is a model of $\Gmc$ w.r.t $\Omc$} if $\cI$ satisfies each fact in $\Gmc$ and each axiom in $\Omc$. For \dlliteR, a \emph{canonical model} exists that can be homomorphically mapped into any other model, obtained from the \emph{deductive closure} $\Omc^\infty(\Gmc)$, which extends $\Gmc$ with triples derived from existing triples in $\Gmc$ by  exhaustively applying axioms in $\Omc$ \cite{DBLP:journals/jar/CalvaneseGLLR07}.

\paragraph{Ontology-Mediated Query Answering} 
A \emph{query atom} is an expression of the form $p(T_1,T_2)$, where $p \in \bR$, and each $T_i \in \mathbf{V} \cup \bI$ is called a \emph{term}, with $\mathbf{V}$ disjoint with $\bI, \bC,$ and $\bR$ being a set of variables. A \emph{monadic conjunctive query} (CQ) $q(X)$ is a First-Order (FO) formula of the form
$\begin{aligned}
    q(X)\leftarrow \exists \vec{Y}. p_1(\vec{T_1}) \land \dots \land p_n(\vec{T_{n}})
\end{aligned}$ where each $p_i(\vec{T_i})$
is a query atom, 
and 
$\vars(q) = \{X\} \cup \vec{Y}$ denotes the set of variables appearing in $q$, with 
$X \not\in \vec{Y}$ being the \emph{answer variable}. In this work, we focus on \textit{monadic Existential Positive FO} (EPFO) queries, i.e.,
unions of monadic CQs~\cite{DBLP:journals/vldb/DalviS07}. 
For a query $q(X)$ and a KG $\Gmc$, a constant $a$ is an answer of $q$ if a mapping $\pi:\!\mathit{var}(q)\mapsto \bI$ exists, s.t. $q\pi \in \Gmc$ and $\pi(X) = a$; $\mi{q[\Gmc]}$ are the answers of $q$ on $\Gmc$.
   
\emph{Ontology-Mediated Query Answering (OMQA)} concerns answering queries by accounting for both the KG and the accompanying ontology.  Since the model constructed from $\Omc^\infty(\Gmc)$ can be homomorphically mapped to every other model, the deductive closure can be used to evaluate queries~\cite{DBLP:journals/jar/CalvaneseGLLR07}.

\begin{definition}
Given a KG $\Gmc$ and an ontology $\Omc$, an entity $a$ from $\Gmc$ is a \emph{certain answer} of $q(X)$ over $(\Gmc, \Omc)$ if $a$ is an answer to $ \mi{q(X)} $ over $\cO^{\infty}(\Gmc)$. We use $q[\Gmc,\Omc]$ to denote the set of \emph{certain answers of $q$ over $(\Gmc,\Omc)$}. 
\end{definition} 

Let $q$ and $q'$ be two monadic queries over $(\Gmc, \Omc)$,
then $q$ is \emph{contained} in $q'$ w.r.t. $\cO$ if $q[\Gmc,\Omc] \subseteq q'[\Gmc,\Omc]$; we call $q$ a \emph{specialization} of $q'$ (written as $q' \overset{\sf s}{\leadsto} q$), and $q'$ a \emph{generalization} of $q$ (written as $q \overset{\sf g}{\leadsto} q'$).
Query generalizations and specializations can be obtained by exploiting ontology axioms; such process (and result) is referred to as \emph{query rewriting}.

\begin{example}
Consider the KG $\Gmc$ in \cref{fig:kgonto} and the query $\mi{q(X)} \gets$ $ \rel{type}(X,\ent{Professor}) \land \rel{degreeFrom}(X, \ent{mit})$. Since $\ent{mat} \! \in \! q[\Gmc]$, it is a certain answer.
Moreover, according to $\Omc$, $\!\ent{AProfessor}$ is a sub-type of $\ent{Professor}$ and $\rel{degreeFrom}$ is inverse of $\rel{hasAlumnus}$, thus $\ent{bob}$ is also a certain answer.
For $q'(X) {\gets} \rel{type}(X,\ent{AProfessor})\wedge \rel{degreeFrom}(X,\ent{mit})$ it holds that $q \overset{\sf s}{\leadsto} q'$ as $\ent{mat}\not \in q'[\Gmc,\Omc]$.
\end{example}

\paragraph{Embedding-Based Approximate Query Answering} 
Since, in reality, KGs might be missing facts, existing query answering techniques designed for complete data might not compute all answers.
In such settings, one assumes that the given KG $\Gmc$ is a subset of a complete but unobservable KG $\Gmc^i$, and one aims at estimating the likely answers to $q$ over $\Gmc^i$.
E.g., $\Gmc^i$ for the graph $\Gmc$ given in \Cref{fig:kgonto} includes the links denoted by dashed edges.
In practice, to evaluate the accuracy of a considered method, $\mathcal{G}^i$ is typically fixed at the beginning, and $\mathcal{G}$ is created by removing facts from $\mathcal{G}^i$. 

The set of all answers to a given query $q$ comprises those that can be obtained by directly querying the given KG $\Gmc$ only and those that require predicting missing KG facts.
Thus, one typically distinguishes easy and hard answers as follows: 
\begin{definition}\label{def:hardans}
Given a KG $\Gmc$, a subgraph of a complete but unobservable KG $\Gmc^i$ and a query $q(X)$, $a$ is an \emph{easy answer} to $q$ if $a \in q[\Gmc]$, and it is a \emph{hard answer} to $q$ if $a \in q[\Gmc^i] \backslash q[\Gmc]$.
\end{definition}
Recently, embedding-based methods have been proposed
for \emph{approximate answering of  existential positive FO queries} over incomplete KGs~\cite{iclr/RenHL20,DBLP:conf/nips/RenL20,DBLP:conf/kdd/LiuDJZT21,DBLP:conf/www/ChoudharyRKSR21,DBLP:conf/aaai/KotnisLN21}. 
Broadly, such methods can be divided into two categories: \emph{query-based}~\cite{iclr/RenHL20,DBLP:conf/nips/RenL20,DBLP:conf/kdd/LiuDJZT21,DBLP:conf/www/ChoudharyRKSR21,DBLP:conf/aaai/KotnisLN21} and \emph{atom-based}~\cite{DBLP:journals/corr/abs-2011-03459}.
Generally, any neural QA model relies on an embedding function which maps entities and relations into a $d$-dimensional embedding space.
It then computes a score of each entity $c$ for being an answer to a given query $q$ over $\Gmc^i$ via a scoring function $\phi_q(\cbf): \mathbb{R}^d \mapsto [0,1]$, where $\cbf$ denotes the embedding vector of $c$.\footnote{Bold small letters denote vector representations.} Using these scoring functions, the final \emph{embedding QA function} $\mathcal{E}_\Gmc$ takes as input a query and returns its approximate answers over the knowledge graph $\Gmc^i$, i.e., answers that have the scoring above some predefined threshold.
We say that $\mathcal{E}_\Gmc$ is \textit{reliable w.r.t. $\Gmc^i$} whenever for each query $q$, $c$ is an approximate answer to $q$ iff $c$ is an answer to $q$ over $\Gmc^i$.
Clearly, the challenge of identifying hard answers is still valid also for embedding QA models.

\begin{table*}[t]

\scriptsize
\begin{center}
	\caption{Rules to specialize and generalize an atom $\beta$ from $q(X) \leftarrow \alpha \land \beta$, where $A,B \in \bC$, $p,r,s \in \bR$ and $T,T_1,T_2 \in \mathit{vars}(q) \cup \bI$. The operators $\overset{\bf s}{\leadsto}$ and $\overset{\bf g}{\leadsto}$ are used for constructing specializations and generalizations respectively of a given query. \label{tab:refrules}}
\resizebox{0.9\textwidth}{!}{%
	\begin{tabular}{l@{If \ }l@{ \quad then: \quad }l@{\quad}l} 
			\textbf{(R1)}\label{R1}	& $A \sqsubseteq B \in {\Omc}\;\;\;\;\;\;$\quad & 
			$\alpha \land \rel{type}(T,B) \overset{\bf s}{\leadsto} \alpha \land \rel{type}(T,A)$ &
			$\;\;\;\;\;\alpha \land \rel{type}(T,A) \overset{\bf g}{\leadsto} \alpha \land  \rel{type}(T,B)$ 
			\\ 
			
			\textbf{(R2)}\label{R2}	& $\exists p\sqsubseteq A  \in {\Omc}$ & 
			$\alpha \land \rel{type}(T_1,A)  \overset{\bf s}{\leadsto} \alpha \land p(T_1,T_2)$ &
			$\;\;\;\;\;\alpha \land p(T_1,T_2) \overset{\bf g}{\leadsto} \alpha \land \rel{type}(T_1,A)$
			\\ 
			
			\textbf{(R3)}\label{R3}	& $A \sqsubseteq \exists p \in {\Omc}$ & 
			$\alpha \land p(T_1,T_2)  \overset{\bf s}{\leadsto} \alpha \land \rel{type}(T,A)$ &
			$\;\;\;\;\;\alpha \land \rel{type}(T_1,A) \overset{\bf g}{\leadsto} \alpha \land p(T_1,T_2)$
			\\ 
			\textbf{(R4)}\label{R4}	& $\exists p^- \sqsubseteq A  \in {\Omc}$ & 
			$\alpha \land A(T)  \overset{\bf s}{\leadsto} \alpha \land p(T_2,T_1)$ &
			$\;\;\;\;\;\alpha \land p(T_2,T_1) \overset{\bf g}{\leadsto} \alpha \land \rel{type}(T_1,A)$
			\\ 
			
			\textbf{(R5)}\label{R5}	& $A \sqsubseteq \exists p  \in {\Omc}$ & 
			$\alpha \land p(T_1,T_2) 
			\overset{\bf s}{\leadsto} \alpha \land \rel{type}(T,A)$ &
			$\;\;\;\;\;\alpha \land \rel{type}(T_1,A) \overset{\bf g}{\leadsto} \alpha \land p(T_1,T_2)$ 
			\\

			\textbf{(R6)} \label{R6}& $p \sqsubseteq s \in {\Omc}$  & 
			$\alpha \land s(T_1,T_2)  \overset{\bf s}{\leadsto} \alpha \land p(T_1,T_2)$ &
			$\;\;\;\;\;\alpha \land p(T_1,T_2) \overset{\bf g}{\leadsto} \alpha \land s(T_1,T_2)$ 
			\\ 
			
			\textbf{(R7)}\label{R7} &	$s^- \sqsubseteq p \in {\Omc}$  & 
			$\alpha \land p(T_1,T_2)  \overset{\bf s}{\leadsto} \alpha \land s(T_2,T_1)$ &
			$\;\;\;\;\;\alpha \land s(T_1,T_2) \overset{\bf g}{\leadsto} \alpha \land p(T_2,T_1)$ 
			\\ 
			
			
			\textbf{(R8)}\label{R8} &	$ \begin{aligned}[t] \theta{:} \mathit{vars}(q) {\mapsto} \mathit{vars}(q) {\cup} \bI \\
			s.t. \ \theta(T_i)=\theta(T_i') \end{aligned} $ & \multicolumn{2}{c}{
				\begin{tabular}[t]{rcl}
					$\alpha \land p(T_1,T_2) \land p(T_1',T_2') \in q$ & $ \overset{\bf s}{\leadsto}$ & $ \alpha\theta \land p(T_1,T_2)\theta $ \\
					$\alpha \land p(T_1,T_2) $ s.t. $T_1 \text{ or }T_2 \in \bI$  & $\overset{\bf g}{\leadsto}$ & $\alpha \land p(Z,T_2)$ or $\alpha \land p(T_1,Z)$
			\end{tabular} } 
		\end{tabular}%
		}
	
	\end{center}
			
	\end{table*}
\section{Embedding-Based OMQA} 
Existing methods for embedding-based query answering compute approximate answers to queries over an unobservable KG $\Gmc^i$ by performing inductive reasoning. However, they are not capable of simultaneously applying deductive reasoning, and thus cannot account for ontologies with which KGs are often accompanied. 

To address this shortcoming, we propose ways to combine inductive and deductive reasoning for approximate query answering over incomplete KGs. For that, we first formalize the task of \emph{Embedding-based Ontology-Mediated Query Answering (E-OMQA)} in which both types of reasoning are exploited. The goal of this task is to approximate certain answers to OMQs over $\Gmc^i$.

\begin{definition}[E-OMQA]\label{def:eomqa}
Let $\Gmc$ be a KG, which is a subgraph of a complete but not observable KG $\Gmc^i$, let $\Omc$ be an ontology and $q$ a query. \emph{Embedding-based ontology-mediated query answering} is concerned with constructing an embedding function $\mathcal{E}_{\mathcal{G,O}}$ that is reliable w.r.t. $\Omc^\infty(\Gmc^i)$.
\end{definition}

\noindent Note that, $q[\Gmc^i,\Omc]$ subsumes both $q[\Gmc^i]$, the answers requiring inductive reasoning, and $q[\Gmc,\Omc]$, the answers computed via deductive reasoning only.
Analogously as for embedding-based query answering, for E-OMQA, we distinguish between easy certain answers and hard certain answers as follows. 

\begin{definition}\label{def:hardcertainans}
Given a KG $\Gmc$, a subgraph of a complete but unobservable KG $\Gmc^i$, an ontology $\Omc$ and a query $q(X)$, $a$ is an \emph{easy certain answer} to $q$ if $a \in q[\Gmc,\Omc]$, and it is a \emph{hard certain answer} to $q$ if $a \in q[\Gmc^i,\Omc] \backslash q[\Gmc,\Omc]$.
\end{definition}

Next, we discuss 
several embedding-based methods for ontology-mediated query answering under incompleteness.

\paragraph{Query Rewriting over Pre-trained Models}
In the traditional OMQA setting, each query $q$ can be evaluated by first rewriting $q$ into a set of FO-queries $Q_\Omc$ and then evaluating each query in $Q_\Omc$ over $\Gmc$ alone. 
For E-OMQA, this amounts to constructing an embedding QA function $\cE_{\Gmc}$ for $\Gmc$ alone and using it to compute the answers to all queries in $Q_{\Omc}$. For \dlliteR, such FO-rewriting is obtained by extensively applying 
ontology axioms in a specializing fashion, which results in the so-called \textit{perfect reformulation}~\cite{DBLP:journals/jar/CalvaneseGLLR07}.

\paragraph{Ontology-Aware Models} 
An alternative to query rewriting is to develop an embedding query answering function 
that accounts for axioms in $\Omc$.
To the best of our knowledge, there are no KGE models that directly address the problem of E-OMQA. Thus, we suggest the following:
(1) Train existing embedding models for 
    QA on the data derived from $\cO^{\infty}(\Gmc)$ instead of $\Gmc$; 
(2) Develop an \emph{ontology-aware} embedding model that will be trained on $\Gmc$ but will have special terms in the training objective 
structurally enforcing $\Omc$. (3) Combine (1) and (2), i.e., train ontology-aware embedding models on the data derived from $\cO^{\infty}(\Gmc)$. 

While the proposed approaches can be
realized on top of any embedding model for logical query answering, in this work, we verify their effectiveness on a query-based model \emph{Query2Box}~\cite{iclr/RenHL20} and an atom-based model \emph{CQD}~\cite{DBLP:journals/corr/abs-2011-03459}. 
In Section~\ref{sec:trainstr}, we present several effective ontology-driven training methods for realizing (1). 
As for (2), building on \emph{Query2Box}, in Section~\ref{sec:trainobj} we develop its ontology-aware version.
Moreover, we build an ontology-aware extension of \cqd, on top of the neural link predictor using \textit{adversarial sets regularization} (ASR)~\cite{DBLP:conf/uai/MinerviniDRR17} to enforce the ontology axioms.
We chose this approach, since it is general and allows us to incorporate rules into any off-the-shelf neural link predictor.
In our experiments, we use ComplEx-N3~\cite{DBLP:conf/icml/LacroixUO18} as it requires minimal modification to CQD and outperforms other neural link predictors (see \cite{DBLP:conf/icml/LacroixUO18}).
Finally, we evaluate the effectiveness of our ontology-driven strategies from Section~\ref{sec:trainstr} on the extended models described in Section~\ref{sec:trainobj} and verify the feasibility of the classical query-rewriting approach in the knowledge graph embedding setting.

\subsection{Ontology-Driven Data Sampling}
\label{sec:trainstr}\label{sec:method}

Let $\Qmc_{\Gmc}$ be the set of all possible queries that can be formed using the signature $\Sigma$. 
During the training process, existing embedding models 
are trained on a set of sampled queries of certain shapes and their answers over the KG $\Gmc$.
%
%
%

\subsubsection{Random Query Sampling}
The existing  
sampling procedure from the literature \cite{nips/HamiltonBZJL18,iclr/RenHL20} arbitrarily chooses entities and relations in the graph to construct queries of various shapes.
Query2Box is trained on complex queries involving multiple atoms, while CQD is trained only on atomic queries, as it relies on a neural link predictor. 
For verifying how well the model generalizes, the test set includes queries whose shapes have not been encountered during training.
Naturally, this procedure is not guaranteed to capture ontological knowledge that comes with the knowledge graph, and using it over $\Omc^\infty(\Gmc)$ could generate a bias towards concepts and roles that are very general.
Moreover, using all possible queries from $\Qmc_{\Gmc}$ with their certain answers might be infeasible in practice.
In the following, we discuss various options for guiding the sampling of queries to train ontology-aware knowledge graph embedding models for query answering.
\begin{figure*}[t] \centering
\includegraphics[width=0.8\textwidth]{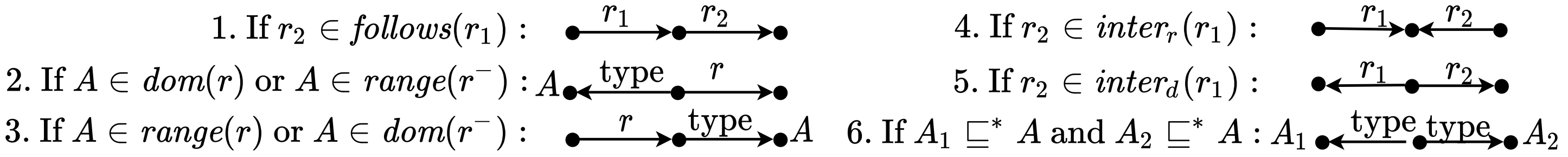}
\caption{Ontology-driven rules to label query shapes; $r^-$ denotes any of $\mathit{inv}(r)$.}
\label{fig:q_onto_shapes}
\end{figure*}

\subsubsection{Incorporating Query Rewritings and Certain Answers} The first natural attempt to incorporate ontologies is to consider certain answers, which for \dlliteR can be done efficiently.
An example of this training case is to randomly sample query $q(Y)\leftarrow\exists X. \rel{hasAlumnus}(\ent{mit}, X) \wedge \rel{worksFor}(X,Y)$ and, given ($\Gmc$, $\Omc$) in Fig.~\ref{fig:kgonto}, use it along with all its certain answers: $\ent{mit, yale}$ during training.
To incorporate the ontology, we can randomly sample queries over the KG, using the standard procedure, and then add their generalizations and 
specializations obtained using the rules in Tab.~\ref{tab:refrules}.
To rewrite a query we select an atom and apply an ontology axiom. For example, the first rule \textbf{(R1)} applies a concept inclusion axiom, while \textbf{(R6)} applies a role inclusion.

The specializations of a query $q$ (\ie $\mi{Spec(q)}$), incorporate more specific information regarding the answers of $q$, while the generalizations of $q$ (\ie $\mi{Gen(q)}$) incorporate additional related entities.
\begin{example} Take the queries ${q_1(X) \!\! \gets \!\! \exists Y. \rel{type}(X, \ent{University})}$ and ${q_2(X) \!\! \gets \!\! \exists Z.\rel{teachesAt}(Z, X)}$.
Using $\textbf{R2}$  in \cref{tab:refrules} and (5) in \cref{fig:kgonto} we get $q_1\! \overset{\sf s}{\!\leadsto\!} q_2$, \ie,  $q_2$ is a specialization of $q_1$. 
\end{example}
\noindent In general, there are exponentially many rewritings, thus we fix a rewriting depth, up to which the respective training queries are generated, via a dedicated parameter. 

\subsubsection{Strategic Ontology-Based Sampling}
While adding generalizations and specializations of randomly selected queries should partially reflect the background knowledge, many relevant axioms can be overlooked if they are not explicitly 
captured in the data.
To overcome this, we consider the set of target
query shapes as directed acyclic graphs (DAGs)
of the form $(N,E)$, where $N$ is a set of nodes and $E \subseteq N \times N$ is a set of directed edges. 
The set of training queries is then obtained by applying a labeling function that assigns symbols in $\Sigma$ to nodes and edges based on the ontology.

\begin{definition}[Query Shape]                   
	A \emph{query shape} $S$ is a tuple $(N,E,n)$ such that $(N,E)$ is a DAG and $n \in N$ is 
	the \emph{distinguished node} of $S$ (\ie, the node for the answer variable). 
	For a given set of relations and constants in $\Sigma$, 
	\emph{a labeling function $f: N \cup E \mapsto\Sigma \cup \Vbf$}
 maps each node to either a variable or an entity and each edge to a relation symbol in the KG signature
	$\Sigma$. 
\end{definition}

\noindent We rely on the ontology when labeling query shapes to create semantically meaningful queries.
 Let $\ISA^*$ be the reflexive and transitive closure of $\ISA$. Then,  for a given relation $p$ we have: 
 \begin{itemize}\small
    \item $\mathit{inv}(p) = \{p' \mid p \sqsubseteq p'^- \in \Omc \}$, \quad $\mathit{dom}(p) = \{ A \mid \exists p'{\sqsubseteq} A' \in \Omc \text{ s.t. } p \ISA^* p', A {\ISA^*} A' \text{ or } A' {\ISA^*} A\}$,
    \item $\mathit{range}(p) {=} \{ A \,{\mid}\, \mi{\exists p'^- {\sqsubseteq} A' \! \in \! \Omc} \text{ s.t. } \mi{p {\ISA^*} p'}, \mi{A  {\ISA^*}\! A'} \!\text{ or }\! \mi{A' {\ISA^*}\! A } \!\}$,
	\item $\mathit{follows}(p) \,{=} \, \{p' \,{\mid} \,{\mathit{range}(p) \cap \mathit{dom}(p')\neq \emptyset} \}$, 
	\item $\mathit{inter}_r(p)=\{p'  \mid  \mathit{range}(p)  \cap  \mathit{range}(p')\neq  \emptyset \text{ or } p_1  \in  \mathit{inv}(p), p_2 \in \mathit{inv}(p') \text{ and }\mathit{dom}(p_1) \cap  \mathit{dom}(p_2)  \neq  \emptyset \}$,
	\item $\mathit{inter}_d(p) =\{p'  \mid  \mathit{dom}(p)  \cap  \mathit{dom}(p') \neq \emptyset \text{ or } p_1  \in  \mathit{inv}(p), p_2  \in  \mathit{inv}(p')$ $\text{ and } \mathit{range}(p_1)  \cap  \mathit{range}(p_2) \neq \emptyset \}$.
 \end{itemize}

Intuitively, for a given relation $p$, the set $\mathit{inv(p)}$ contains all inverse relations of $p$, $\mathit{dom}(p)$ contains all domain types for $p$, $\mathit{range}(p)$ 
all range types for $p$, $\mathit{follows}(p)$ 
stores all relations $p'$ which can follow $p$, and $\mathit{inter}_r(p)$, $\mathit{inter}_d(p)$ contain resp. all relations $p'$ which can intersect with $p$ on range and domain positions. Then, for each shape we label nodes and edges to create queries that are valid w.r.t. $\Omc$ as shown in \cref{fig:q_onto_shapes}.
Note that 
this query sampling process uses only the ontology, \ie, it is data independent. 

\subsection{Ontology-Aware Query2Box}\label{sec:trainobj}
Query2Box~\cite{iclr/RenHL20} is a prominent \emph{query-based} embedding models, in which entities and queries are embedded as \emph{points} and \emph{boxes}, resp., in a $d$-dimensional vector space. 
 A \emph{$d$-dimensional embedding}
 	is a function $\varphi$ that maps $c \in \bI \cup \bC$ to $\cbf \in \mathbb{R}^d$ and a query $q$
 	to $\qbf {=} (\cen_q, \off_q) {\in} \mathbb{R}^{d} \times\mathbb{R}_{\geq 0}^{d}$, which is used to define a \emph{query box} as \begin{center}
     $\text{box}_q = \{\mathbf{v} \in \mathbb{R}^d \mid \cen_q - \off_q \preceq \mathbf{v} \preceq \cen_q + \off_q\},$
 \end{center}
 where $\preceq$ is the element-wise inequality, $\cen_q$ is the center of the box, and $\off_q$ is the positive offset of the box, modeling its size. The score for an entity $c$ being an answer to $q$ is computed based on the distance from $\cbf$ to $\text{box}_q$. The Query2Box model relies on the following geometric operators.

\subsubsection{Projection}
Let $S \subseteq \bI \cup \bC$ be a set of entities, and $r \in \bR$ a relation.
Intuitively, the \emph{projection} operator performs graph traversal, e.g. given an entity $e$, the projection operator for the relation $r$ provides the box corresponding to the set $\{e' \in \bI \cup \bC \mid r(e,e') \in \Gmc\}$. Given the embedding $\mathbf{r} = (\cen_r, \off_r) \in \mathbb{R}^d \times \mathbb{R}^d_{\geq 0}$ for the relation $r$, 
we model the projection of a box $\mathbf{v}=(\cen_v, \off_v)$ by applying element-wise summation $\mathbf{v}+\mathbf{r}=(\cen_v + \cen_r, \off_v + \off_r)$.
This relational translation~\cite{transe} operation corresponds to the translation and enlargement of the box $\mathbf{v}$.

\subsubsection{Intersection}
Given a set of entity sets $\{S_1, \dots, S_n\}$, each of which is represented by a box in Query2Box, 
the \emph{intersection} operator computes their intersection. 
The intersection $\mathbf{w} = (\cen_w, \off_w)$ of a set of boxes $\{(\cen_{v_1},\off_{v_1}), \ldots, (\cen_{v_n},\off_{v_n}) \}$ 
for $\{S_1, \ldots, S_n\}$ is modeled by applying the following operations:
\begin{align*}
    \cen_w &= \sum_{i=1}^n \Phi\big(\text{NN}(\cen_{v_1}), \dots, \text{NN}(\cen_{v_n})\big)_i \odot \cen_{v_i}, \\
    \off_w &= \min(\off_{v_1}, \dots, \off_{v_n}) \odot \sigma\big(\Psi(\off_{v_1}, \dots, \off_{v_n})\big),
\end{align*}
where $\odot$ and $\min$ denote the element-wise multiplication and minimum, respectively. 
$\text{NN} \colon \mathbb{R}^d \to \mathbb{R}^d$ is a 2-layer feed-forward neural network having the same dimensionality for the hidden layers as for the input layer.
$\Phi$ and $\sigma$ stand for the softmax and sigmoid functions, resp., applied in a dimension-wise manner.
$\Psi$ is a permutation invariant function composed of a 2-layer feed-forward network followed by element-wise mean operation and a linear transformation.
The center $\cen_w$ is calculated as the weighted mean of the box centers $\cen_{v_1}, \dots, \cen_{v_n}$.
This geometric intersection provides a smaller box that lies inside a given set of boxes -- for more details, we refer the reader to~\cite{iclr/RenHL20}.

The goal of the Query2Box model is to learn the embedding of queries, such that the \emph{distance} between the box corresponding to the query and its answers is minimized, while the \emph{distance} to this box from other randomly sampled non-answers is maximized.

In what follows, we present our proposal for integrating ontological axioms into the Query2Box model. Similarly to \cite{iclr/RenHL20}, we define the distance between 
$\mathbf{q} \in \mathbb{R}^d \times \mathbb{R}_{\geq 0}^d$ and $\mathbf{v}\in \mathbb{R}^d$ as $d(\mathbf{q}, \mathbf{v}) = \| \cen_q - \mathbf{v}\|_1$, namely the $L_1$ distance from the entity $\mathbf{v}$ to the center of the box.
Using the sigmoid function we transform the distance into the $(0,1)$ interval, that is, $p(\mathbf{v}\, |\, \mathbf{q}) = \sigma\big(-(d(\mathbf{q}, \mathbf{v}) - \gamma)\big)$, where $\gamma > 0$ is a margin, which
denotes the probability of 
$v \in q[\Gmc^i,\Omc]$.

\begin{figure}[t] \centering
        \includegraphics[width=.47\textwidth]{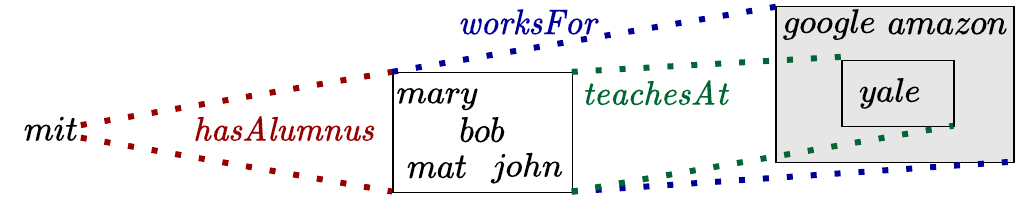}
    \caption{
   Illustration of our extension of Query2Box. The KG nodes and relations are embedded as points and projection operators, resp. 
    The axiom $\rel{teachesAt} \ISA \rel{worksFor}$ is captured by the inclusion of the respective boxes for queries.
    \label{fig:boxes}} 
\end{figure}
Note that for every ontological axiom its both left- and right-hand side can be turned into queries.
When embedding those queries as boxes, axioms can be naturally enforced 
if in the vector space the inclusion of the boxes corresponding to the respective queries is ensured. For a query $q$, let $Gen(q) = \{q_1\dotsc q_n\}$ be the set of all generalizations of $q$ based on $\Omc$.
Given a train query $q$ and $v \in q[\Gmc, \Omc]$, we aim at maximizing $\prod_{i=1}^n p(\mathbf{v}\, |\, \mathbf{q}_i)^{\beta_i}$, where $\beta_i \geq 0$ is a weighting parameter for all $i=1, \dots, n$.
This is achieved by minimizing the negative log-likelihood:
$
    -\log\Big(\prod_{i=1}^n p(\mathbf{v}\, |\, \mathbf{q}_i)^{\beta_i}\Big) = -\sum_{i=1}^n \beta_i \log\big(p(\mathbf{v}\, |\, \mathbf{q}_i)\big).
$
By exploiting that $\sigma(x) = 1 - \sigma(-x)$, for any $\mathbf{v}'_j \not \in q[\Gmc, \Omc]$, we have 
$
    p(\mathbf{v}'\, |\, \mathbf{q}) = 1 - p(\mathbf{v}\, |\, \mathbf{q}_i) = \sigma(d(\mathbf{q}, \mathbf{v}) - \gamma) \; .
$

Our goal is to enforce that if $q' \in \mathit{Gen}(q)$ 
then the box of $q'$ 
contains 
the box of $q$. 
In order for that to hold, we need to ensure that, if $a$ is an answer to $q$ then the distance not only between $a$ and $q$ should be minimized, but also between $a$ 
and all 
generalizations 
of $q$. The following training objective reflects our goal:
\begin{equation*}
L\!=\!-\sum_{i=1}^n \beta_i \log\sigma\big(\gamma - d(\mathbf{v},\mathbf{q}_i)\big)
-\sum_{j=1}^k \frac{1}{k} \log\sigma(d(\mathbf{v}'_j; \mathbf{q}) - \gamma),
\end{equation*} 
where $\mathbf{v}'_j \not \in q[\Gmc, \Omc]$ is a random entity for all $j=1, \dots, k$ obtained via negative sampling. 
In our experiments, we use $\beta_i=|Gen(q)|^{-1}=\nicefrac{1}{n}$.
\begin{example}
In \cref{fig:boxes}, the entities and relations 
are embedded into the vector space as points and projection operators, resp.
The embedding of 
 $q(Y)\!\! \leftarrow \!\! \exists X.\rel{hasAlumnus}(\ent{mit}, X)\! \wedge\!\rel{worksFor}(X, Y)$
is represented by the larger gray box, obtained by applying the projection $\rel{hasAlumnus}$ to the embedding of entity $\ent{mit}$ followed by the projection on $\rel{worksFor}$.
To enforce $\rel{teachesAt \!\sqsubseteq\! worksFor}$ we ensure that the box of $q'(Y) \leftarrow \exists X.\rel{hasAlumnus}(\ent{mit}, X)\land \rel{teachesAt}(X, Y)$, is contained in the box corresponding to $q$.
 \end{example}

Conceptually, our training data sampling techniques and the loss function modifications are flexible in terms of the DL, in which the ontology is encoded. The only restriction is the existence of efficient query rewriting 
algorithms.

\subsection{Ontology-aware CQD}
A prominent 
 atom-based query-answering method is CQD \cite{DBLP:journals/corr/abs-2011-03459}, which
 relies on neural link predictors for answering atomic sub-queries, and then aggregates the resulting scores via t-norms. 
 
We now describe how we inject the ontology axioms into the neural link predictor employed by  CQD~\cite{DBLP:journals/corr/abs-2011-03459}. For that we rely on the FO translation of the DL axioms.  Following \cite{DBLP:conf/uai/MinerviniDRR17}, for each rule the goal is to identify the entity embeddings which maximize an \emph{inconsistency loss}, i.e., the entities for which the scoring of the head is much lower compared to the scoring of the body. For example, given the rule $\Gamma$:  $\forall X,Y~\rel{teachesAt}(X,Y) \rightarrow \rel{type}(Y,\rel{University})$, the goal is to map the variables to $d$-dimensional embeddings, \ie $\phi: \mathit{var}(\Gamma) \mapsto \mathbb{R}^d$, s.t. $[\mathit{score}_{\rel{teachesAt}}(\phi(X),\phi(Y)) - \mathit{score}_{\rel{type}}(\phi(Y), \mathbf{University})]_+$ is maximal, where $[x]_+ = \mathit{max}([x],0)$ with $[x]$ being the integral part of x, and $\mathit{score}_r$ being the scoring function for the relation $r$ determining 
whether there is an $r$-edge between any two given entities. Mapping $\phi$ determines a so-called \emph{adversarial input set}, which is used as an adaptive regulariser for the neural link predictor.  The inconsistency loss is then incorporated into the final loss function of the ontology-aware model which tries to minimize the maximal inconsistency loss while learning to predict the target graph over the given sets of correct triples. In experiments we rely on the existing implementation of the adversarial sets regularisation method in ComplEx-N3, which is the default neural link predictor for CQD.

 \begin{table*}[t] \small
     \caption{The total number of axioms $|\Omc|$ and of each type, the size of the input KG $|\Gmc|$, the number of entities \textbf{$|\bI|$}, the number of relations \textbf{$|\bR|$}, and the number of materialized triples \textbf{$|\Omc^\infty(\Gmc)|$}.}
    \label{tab:OntStatistics}
    \centering
    \begin{tabular}{c|cccccc | cccc}
    \toprule
    \multirow{2}{*}{\textbf{Dataset}}& \multicolumn{6}{c}{\textbf{Ontology \Omc}} & \multicolumn{4}{c}{ \textbf{KG $\Gmc$}} \\
          &  $|\Omc|$ & $A \ISA A'$ & $p \ISA s$ & $p^- \ISA s$ & $\exists p \ISA A$ & $\exists p^- \ISA A$ & $|\Gmc|$ & \textbf{$|\bI|$} & \textbf{$|\bR|$} & \textbf{$|\Omc^\infty(\Gmc)|$}\\ 
          
          \hline
          
         LUBM & 68 & 13 & 5& 28 & 11 & 11 &284K & 55684 & 28 & 565K\\
         NELL & 307 & -- & 92 & 215 & -- & -- &285K & 63361 & 400 & 497K\\
    \bottomrule
    \end{tabular}
 
\end{table*}

\begin{table*}[t] \small
    \centering
     \caption{Number of test and train queries of each shape in each of the settings.}
    \label{tab:Queries}
    \begin{tabular}{c|c|ccccccccc} \toprule
         \multirow{2}{*}{\textbf{Dataset}} & \multirow{2}{*}{\textbf{Train/Test}} & \multicolumn{9}{c}{\textbf{Query Shape}}  \\
         & & 1p & 2p & 3p & 2i & 3i & ip & pi & 2u & up \\ 
         \hline
         \multirow{7}{*}{LUBM} & \plain & 110000& 110000 &110000 & 110000 & 110000 & -- & -- & -- & -- \\
                            &   \gen  & 117124 & 136731 & 150653 & 181234 & 208710 &  -- & -- & -- & -- \\
                            & \spe & 117780 & 154851 & 173678 & 271532 & 230085 & -- & -- & -- & --\\
                            & \onto & 116893 & 166159 & 333406 & 212718 & 491707 & -- & -- & -- & -- \\ \cline{2-11}
                            & \tstA &  8000 & 8000 & 8000 & 8000 & 8000 & 8000 & 8000 & 8000 & 8000\\
                            & \tstB & 1241 & 4701 & 6472 & 3829 & 4746 & 7393 & 7557& 4986 & 7122\\
                            & \tstC & 8000 & 8000 & 8000 & 8000 & 8000 & 8000 & 8000 & 7986 & 8000\\ \hline
         \multirow{7}{*}{NELL} & \plain & 107982 & 107982 & 107982 & 107982 & 107982 &  -- & -- & -- & -- \\
                            &   \gen  & 174310 & 408842 & 864268 & 398412 & 930787 &  -- & -- & -- & -- \\
                            & \spe & 174310 & 419664 & 906609 & 401954 & 936537 &  -- & -- & -- & --  \\
                            & \onto & 114614 & 542923 & 864268 & 629144 & 930787 & -- & -- & -- & --  \\ \cline{2-11}
                            & \tstA & 15688 & 3910 & 3918 & 3828 & 3786 & 3932 & 3895 & 3940 & 3966 \\
                            & \tstB & 346 & 4461 & 4294 & 4842 & 5996 & 7295 & 5862 & 5646 & 6894\\
                            & \tstC & 8000 & 8000 & 8000 & 8000 & 8000 & 8000 & 8000 & 7990 & 8000 \\ \bottomrule
    \end{tabular}

\end{table*}

 \begin{figure*}[t] \centering
\includegraphics[scale=.5]{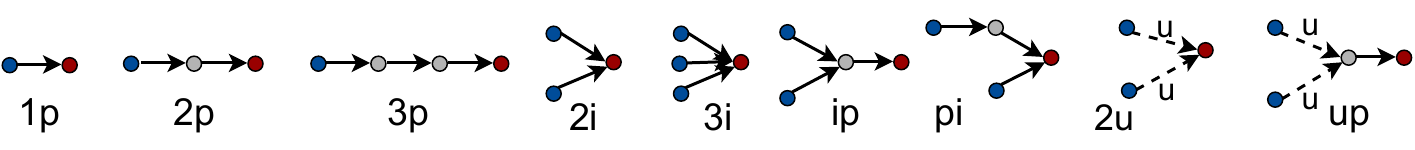}
\caption{Query shapes considered in our experiments, where blue nodes correspond to anchor entities and red ones to answer variables; p stands for projection, i for intersection and u for union. The first five shapes are used in training.}
\label{fig:q_shapes} 
\end{figure*}
\section{Evaluation}\label{sec:eval}

We evaluate our
 training strategies on 
 popular models: 
 Query2Box (\qbox) 
 and \cqd, as well as our ontology-aware adaptations
\obox and \cqdasr. 
 Specifically, we test their 
ability to perform inductive reasoning, deductive reasoning, and their combination. 

\subsection{Benchmarks for E-OMQA}

Since the task of embedding-based ontology mediated query answering has not been considered in the literature before, no benchmarks for it existed prior to our work. 
Thus, we have created two novel benchmarks based on LUBM \cite{lubm} and NELL \cite{nell} KGs,
available online\footnote{\url{https://github.com/medinaandresel/eomqa}}. LUBM has a rich ontology including 
domain and range axioms as well as concept and role inclusions, while the NELL KG is accompanied with a more simple  ontology containing only (inverse) role inclusions.
Following common practice, each input KG is completed w.r.t. inverse edges. 
In \cref{tab:OntStatistics} we present the number of ontology axioms of various types as well as the number of (materialized) triples, entities and relations in these datasets.

\subsubsection{Query and Answers Sampling} We use the same type of queries (corresponding to directed acyclic graphs with entities as the source nodes, also known as \emph{anchors}) as in prior work~\cite{iclr/RenHL20} (see \Cref{fig:q_shapes}). 
We assume that each input KG 
is 
complete (\ie $\Gmc^i$) and then 
partition 
it into $\Gmc_{valid}$ for validation and $\Gmc_{train}$ for training by discarding 10\% of edges at each step; this yields 
$\Gmc_{train} \subsetneq \Gmc_{valid} \subsetneq \Gmc$. 
We then create several training sets of queries according to 
our ontology-aware data sampling strategies 
from Section~\ref{sec:trainstr}: 

	\noindent$\bullet$ $\mi{plain}$: 
	the training queries are randomly sampled 
	from 
    $\Gmc_{\mi{train}}$, and 
	we take their plain answers, \ie over $\Gmc_\mi{train}$.
 
	\noindent $\bullet$ $\mi{gen}$: 
	queries  
	in $\mi{plain}$ augmented with their
	ontology-based generalizations\footnote{This is similar to random sampling over $\Omc^\infty(\Gmc_\mi{train})$ but unlike the deductive closure, our procedure is guaranteed to terminate. We used the rewriting depth of up to 10.}; 
	answers are certain, \ie, over $\Omc^\infty(\Gmc_\mi{train})$. 

	\noindent $\bullet$ $\mi{spec}$: queries from $\mi{gen}$ augmented with specializations; 
 	
 	\noindent $\bullet$ $\mi{onto}$: 
 	queries from  
 	\cref{sec:trainstr}, 
with randomly chosen 
percentage of 
valid entities as anchors; all answers are certain. 

Specializations and generalizations non-compliant with the shapes from Figure~\ref{fig:q_shapes} are discarded.
Note that for NELL, the $\mathit{plain}$ data is exactly the one from \cite{DBLP:conf/nips/RenL20}. We observe that the number of 1p queries obtained for $\mathit{gen}$ and $\mathit{spe}$ settings are identical. This is probably because 
the set of 1p queries in $\mathit{plain}$ covers all edges in the training KG.
For the LUBM dataset, we have created the training and testing sets from scratch, and the 1p queries in $\mathit{plain}$ do not contain the entire training KG. The $\mi{onto}$ set of queries leverages the proposed ontology-driven technique, given that the ontology covers all relations and concepts in the KG and describes how they interact, \ie, the ontology axioms support all the constructed queries, and we chose 50 \% of valid entities as anchors.  As there are too many queries to chose from, due to the large number of relations, we had to select a smaller number of valid entities as anchors, namely 20-30\%.
 This explains the smaller number of 1p queries. Moreover, the NELL ontology does not contain interesting axioms that can be leveraged by ontology-driven query sampling technique, thus to obtain $\mi{onto}$ we had to rely on the patterns from the data alone. 
\normalsize

We generate three different test sets for verifying the ability of the query answering model to perform inductive reasoning, deductive reasoning and their combination. More formally, 

\begin{itemize} 
	\item \textbf{Inductive case (I)}.
	Is the model able to predict missing answers to queries over the  complete, but not observable KG $\Gmc^i$? (accounts for the standard test case)
	\item \textbf{Deductive case (D)}. 
Is the model able to predict answers that can be inferred from the known triples in $\Gmc_{\mi{train}}$ using ontology?
\item \textbf{Inductive + Deductive case (I+D)}. 
	Is the model able to predict missing answers 
	inferred from
 the complete but not observable KG $\Gmc^i$
using axioms in $\Omc$?
\end{itemize}
\smallskip

For test case \tstA, respectively \tstC, test queries are randomly sampled over $\Gmc$, respectively $\Omc^{\infty}(\Gmc)$, while for 
\tstB\ they are randomly sampled  over $\Omc^{\infty}(\Gmc_\mathit{train})$ s.t. they 
cannot be trivially answered over $\Gmc_\mi{train}$, and unseen during training. In each test case the validation queries are generated similarly but over $\Gmc_{valid}$. 

The size of each training/testing set, and the number of queries per shape for each of the considered cases are presented in \cref{tab:Queries}. 

For each test and validation query, we measure the accuracy based on \emph{hard (certain) answers}, \ie, those that  cannot be trivially answered over $\Gmc_\mi{train}$ (or $\Gmc_\mi{valid}$ for test queries) and require 
prediction of missing edges and/or application of ontology axioms (see Definition~\ref{def:hardans} and~\ref{def:hardcertainans}).

\subsection{Models and Evaluation Procedure}
 We consider  \qbox, \obox, \cqd and \cqdasr trained in each described setting: 
\ie, $M_x$, where $M \in \{$\qbox, \obox, \cqd, \cqdasr$\}$ and $x \in \{$plain, gen, spec, onto$\}$; 
\qbplain and \cqdplain\ are taken as baselines. \qbox and \obox are trained on five query shapes that require projection and intersection, while \cqd and \cqdasr are trained on atomic queries. 
 We have configured both \qbox and \obox systems as follows:
The size of the embedding dimension was set to $400$, and the models were trained for $15 \times 10^{4}$ steps 
using Adam optimizer
with an initial learning rate of $10^{-4}$ and the batch size of 512.   The rest of the parameters were set in the same way as in \cite{iclr/RenHL20}.

For CQD, we used the code 
from \cite{DBLP:journals/corr/abs-2011-03459} with ComplEx-N3~\cite{DBLP:conf/icml/LacroixUO18} employed as the base model. The embedding size was set to 1000, and the regularisation weight was selected based on the validation set by searching in $\{ 10^{-3}, 5 \times 10^{-3}, \ldots, 10^{-1} \}$.  For LUBM, the regularization weight was set to $0.1$ in the $\mathit{gen}$, $\mathit{spe}$, and $\mathit{onto}$ settings, and to $0.01$ in the $\mathit{plain}$ setting.
For NELL, the regularization weight was set to $0.005$ in the $\mathit{plain}$ setting, to $0.001$ in the $\mathit{gen}$ and $\mathit{spe}$ settings, and to $0.05$ in the $\mathit{onto}$ setting.
For \cqdasr we have additionally used a regularisation weight of the following values: $10^{-2}$, $10^{-3}$ and $10^{-4}$. The batch size of the adversarial examples was set to 32. 

We evaluated the models periodically and 
report the test results of the models with 
the best performance on the validation dataset.  
The performance of each trained model  
is measured using standard metric HITS@$K$ for $K$=3 (HITS@$3$), 
indicating the frequency that the correct answer is ranked among the top-$3$ results.


\begin{figure*}[t!]\centering
\captionsetup[subfigure]{labelformat=empty}
\resizebox{\linewidth}{!}{\centering \begin{subfigure}[t]{0.8\textwidth} \centering
\pgfplotsset{ytick style={draw=none}, 
  legend columns=2, 
  legend style={
      draw=,/tikz/every even column/.append style={column sep=0.3cm},
      fill,
      at={(0.1,-0.1)},
      legend columns=2,
      legend cell align=left,
      anchor=north west
      },
  legend cell align={left}}
\begin{tikzpicture}
\begin{axis}[
   axis y line*=right,
   symbolic x coords={plain,gen,spe,onto},
   ybar,
   xtick align=inside,
   xtick=data,
   ymajorgrids,
   bar width=0.3cm,
   ylabel style={align=center},
   ylabel={\#Training Queries},
   nodes near coords style={font=\tiny},
   ylabel near ticks, yticklabel pos=right
   ]
   \addplot[ybar,ybar legend,fill=gray] table [x=setup, y=queries] {q2b_data_lubm.csv};\label{Q2Bdataplot}               
    \addplot[ybar,ybar legend,fill=teal!40] table [x=setup, y=queries] {o2b_data_lubm.csv};\label{O2Bdataplot} 
    \addplot[ybar,ybar legend,fill=red] table [x=setup, y=queries] {cqd_data_lubm.csv};\label{CQDdataplot} 
    \addplot[ybar,ybar legend,fill=orange] table [x=setup, y=queries] {cqd_asr_data_lubm.csv};\label{CQDASRdataplot} 
\end{axis}
\begin{axis}[
   axis y line*=left,
   symbolic x coords={plain,gen,spe,onto},
   xtick=data,
   ylabel={HITS@3},
   ylabel near ticks,
   xticklabels={,,},
   nodes near coords style={font=\tiny},
   y tick label style={
        /pgf/number format/.cd,
        fixed,
        fixed zerofill,
        precision=1,
        /tikz/.cd
    }
   ]
   \addplot[black,solid,every mark/.append style={solid,fill=gray},mark=*] table [x=setup, y=performance] {q2b_data_lubm.csv};
   \addlegendentry{Avg. HITS@3 \qbox}
   \addlegendimage{/pgfplots/refstyle=Q2B}
   \addlegendentry{Train \qbox}
   
   \addplot[teal,solid,every mark/.append style={solid,fill=teal!40},mark=square*] table [x=setup, y=performance] {o2b_data_lubm.csv};
   \addlegendentry{Avg. HITS@3 \obox}
   \addlegendimage{/pgfplots/refstyle=O2B}
   \addlegendentry{Train \obox}
   
     \addplot[red,solid,every mark/.append style={solid,fill=red},mark=diamond*] table [x=setup, y=performance] {cqd_data_lubm.csv};
   \addlegendentry{Avg. HITS@3 \cqd}
   \addlegendimage{/pgfplots/refstyle=CQD}
   \addlegendentry{Train \cqd}
   
   \addplot[orange,solid,every mark/.append style={solid,fill=orange},mark=triangle*] table [x=setup, y=performance] {cqd_asr_data_lubm.csv};
   \addlegendentry{Avg. HITS@3 \cqdasr}
   \addlegendimage{/pgfplots/refstyle=CQDASR}
   \addlegendentry{Train \cqdasr}
\end{axis}
\end{tikzpicture}
\caption{LUBM}
\label{lubm:datatstC}
\end{subfigure} 
\hfil %
\begin{subfigure}[t]{0.8\textwidth} \centering
\pgfplotsset{ytick style={draw=none}, 
  legend columns=2, 
  legend style={
      draw=,/tikz/every even column/.append style={column sep=0.3cm},
      fill,
      at={(0.1,-0.1)},
      legend columns=2,
      legend cell align=left,
      anchor=north west
      },
  legend cell align={left}}
\begin{tikzpicture}
\begin{axis}[
   axis y line*=right,
   symbolic x coords={plain,gen,spe,onto},
   ybar,
   xtick align=inside,
   xtick=data,
   ymajorgrids,
   bar width=0.3cm,
   ymin=0, 
   ylabel style={align=center},
   ylabel={\#Training Queries},
   nodes near coords style={font=\tiny},
   ylabel near ticks, yticklabel pos=right
   ]
   
   \addplot[ybar,ybar legend,fill=gray] table [x=setup, y=queries] {q2b_data_nell.csv};\label{Q2B}               
    \addplot[ybar,ybar legend,fill=teal!40] table [x=setup, y=queries] {o2b_data_nell.csv};\label{O2B} 
    \addplot[ybar,ybar legend,fill=red] table [x=setup, y=queries] {cqd_data_nell.csv};\label{CQD} 
    \addplot[ybar,ybar legend,fill=orange] table [x=setup, y=queries] {cqd_asr_data_nell.csv};\label{CQDASR}
\end{axis}
\begin{axis}[
   axis y line*=left,
   symbolic x coords={plain,gen,spe,onto},
   xtick=data,
   ylabel={HITS@3},
   ylabel near ticks,
   xticklabels={,,},
   nodes near coords style={font=\tiny},
   y tick label style={
        /pgf/number format/.cd,
        fixed,
        fixed zerofill,
        precision=1,
        /tikz/.cd
    }
   ]
   \addplot[black,solid,every mark/.append style={solid,fill=gray},mark=*] table [x=setup, y=performance] {q2b_data_nell.csv};
   \addlegendentry{Avg. HITS@3 \qbox}
   \addlegendimage{/pgfplots/refstyle=Q2B}
   \addlegendentry{Train \qbox}
   
   \addplot[teal,solid,every mark/.append style={solid,fill=teal!40},mark=square*] table [x=setup, y=performance] {o2b_data_nell.csv};
   \addlegendentry{Avg. HITS@3 \obox}
   \addlegendimage{/pgfplots/refstyle=O2B}
   \addlegendentry{Train \obox}
   
     \addplot[red,solid,every mark/.append style={solid,fill=red},mark=diamond*] table [x=setup, y=performance] {cqd_data_nell.csv};
   \addlegendentry{Avg. HITS@3 \cqd}
   \addlegendimage{/pgfplots/refstyle=CQD}
   \addlegendentry{Train \cqd}
   
   \addplot[orange,solid,every mark/.append style={solid,fill=orange},mark=triangle*] table [x=setup, y=performance] {cqd_asr_data_nell.csv};
   \addlegendentry{Avg. HITS@3 \cqdasr}
   \addlegendimage{/pgfplots/refstyle=CQDASR}
   \addlegendentry{Train \cqdasr}
\end{axis}
\end{tikzpicture}
\caption{NELL}
\label{nell:datatstC}
\end{subfigure}}
\caption{Performance of \qbox,\obox,\cqd, and \cqdasr on \tstC~ and size of the training set for each setting $\mi{plain}$, $\mi{gen}$, $\mi{spe}$, $\mi{onto}$. The number of training queries is scaled by multiplying with $10^6$.}
\label{fig:dataAug}
\end{figure*}

\begin{savenotes}
\begin{table*}[t]
    \caption{HITS@3 scores in the inductive and deductive setting (\textbf{I+D}) for each query shape (the higher the better)}
    \label{tab:IDcaseLUBMNELL}
    \centering
    \small
    \renewcommand{\arraystretch}{1.1}
    \setlength{\tabcolsep}{2.2pt}
    \resizebox{\textwidth}{!}{%
    \begin{tabular}{l| >{\columncolor{gray!15}}c| ccc| c c| c c| c c ||  >{\columncolor{gray!15}}c| ccc| c c| c c| c c }
    
    \toprule
\textbf{Dataset} &  \multicolumn{10}{c}{\textbf{LUBM}} &  \multicolumn{10}{c}{\textbf{NELL}}   \\
    \toprule
   \textbf{Model}& \textbf{Avg.} & \textbf{1p} &\textbf{2p} & \textbf{3p} & \textbf{2i} & \textbf{3i} & \textbf{ip} &  \textbf{pi} & \textbf{2u} & \textbf{up} & \textbf{Avg.} & \textbf{1p} &\textbf{2p} & \textbf{3p} & \textbf{2i} & \textbf{3i} & \textbf{ip} &  \textbf{pi} & \textbf{2u} & \textbf{up}  \\ 
  \toprule
    
    $\mathit{Q2B_{\mi{plain}}}$ & 
    0.218 & 0.173 &  0.101 & 0.107 & 0.433 & 0.546 & 0.167 & 0.200 & 0.133 & 0.100 &
    0.458 &  0.516 & 0.343 & 0.286 & 0.747 & 0.81 & 0.404 & 0.447 & 0.325 & 0.241 \\
    $\mathit{O2B}_{\mi{plain}}$ & 
   0.245 & 0.235 & 0.109 & 0.095 & 0.488 & 0.584 & 0.176 & 0.218 & 0.2 & 0.103 & 
  \textbf{0.596} & \textbf{0.79} & 0.409 & \textbf{0.359} & \textbf{0.904} & \textbf{0.936} & 0.479 & \textbf{0.521} & \textbf{0.666} & 0.303 
    \\    
    $\mathit{CQD_{\mi{plain}}}$ & 
    0.179 & 0.109 & 0.058 & 0.104 & 0.384 & 0.502 & 0.130 & 0.187 & 0.092 & 0.046 &
    0.555    & 0.664 & 0.383 & 0.304 & 0.853 & 0.903 & 0.471 & 0.512 & 0.599 & 0.306 \\ 
    
    $\mathit{CQD^{\mi{ASR}}_{\mi{plain}}}$ &
    \textbf{0.56} & \textbf{0.682} & \textbf{0.589} & \textbf{0.393} & \textbf{0.659} & \textbf{0.664} & \textbf{0.547} & \textbf{0.488} & \textbf{0.509} & \textbf{0.509} &
    0.592 & 0.716 & \textbf{0.518} & 0.337 & 0.807 & 0.831 & \textbf{0.547} & 0.513 & 0.614 & \textbf{0.445} \\

    \midrule 
    
    $\mathit{Q2B_\mi{gen}}$ & 
0.458   & 0.592 & 0.267 & 0.129 & \textbf{0.789} & \textbf{0.870} & 0.360 & 0.282 & 0.552 & 0.279 & 
0.642    & 0.858 & 0.485 & 0.397 & 0.928 & 0.95 & 0.538 & 0.539 & 0.768& 0.312 \\
    $\mathit{O2B_\mi{gen}}$ & 
0.447    & 0.577 & 0.257 & 0.114 & 0.777 & 0.859 & 0.359 & 0.27 & 0.546 & 0.264 
    &
0.652 & 0.859 & 0.494 & 0.420 & 0.928 & 0.953 & 0.552 & 0.559 & 0.77 & 0.329 \\ 
    
    $\mathit{CQD_\mi{gen}}$ & 
   0.408 & 0.539 & 0.214 & 0.098 & 0.710 & 0.791 & 0.304 & 0.302 & 0.513 & 0.208 & 
  \textbf{0.809}    &\textbf{0.903} & \textbf{0.775} & \textbf{0.473} & \textbf{0.957} & \textbf{0.969} & \textbf{0.821} & \textbf{0.757} & \textbf{0.886} & \textbf{0.743} \\ 

    $\mathit{CQD^{\mi{ASR}}_{\mi{gen}}}$ &  
    \textbf{0.628} & \textbf{0.733} & \textbf{0.640} &\textbf{0.413} & 0.717 & 0.720 & \textbf{0.598} & \textbf{0.599} & \textbf{0.653} & \textbf{0.582} &
    0.787 & 0.9 & 0.771 & 0.467 & 0.919 & 0.924 & 0.793 & 0.723 & 0.846 & 0.741 \\

    \midrule
    
    $\mathit{Q2B_\mi{onto}}$ & 
0.687    & 0.762 & 0.617 & 0.447 & 0.868 & 0.915 & 0.693 & 0.555 & 0.732 & 0.600 &
0.636   & 0.858 & 0.472 & 0.398 & 0.927 & 0.948 & 0.529 & 0.524 & 0.747 & 0.317 
    \\
    $\mathit{O2B_\mi{onto}}$ &  
0.707 & \textbf{0.771} & 0.629 & 0.476 & \textbf{0.878} & \textbf{0.927} & 0.694 & 0.619 & \textbf{0.752} & \textbf{0.618} &
0.655  & 0.862 & 0.498 & 0.423 & \textbf{0.933} & \textbf{0.953} & 0.557 & 0.555 & 0.773 & 0.340 
    \\
 
 $\mathit{CQD_\mi{onto}}$ &  
\textbf{0.723}  & 0.752 & \textbf{0.681} & \textbf{0.481} & 0.870 & 0.924 & \textbf{0.735} & \textbf{0.728} & 0.738 & 0.604 & 
0.545 & 0.667 & 0.368 & 0.293 & 0.848 & 0.904 & 0.453 & 0.506 & 0.595 & 0.275 \\

$\mathit{CQD^{\mi{ASR}}_{\mi{onto}}}$  & 0.664 & 0.753 & \textbf{0.681} & 0.421 & 0.744 & 0.755 & 0.643 & 0.666 & 0.704 & 0.615 &
\textbf{0.77} & \textbf{0.9} & \textbf{0.741} & \textbf{0.456} & 0.922 & 0.923 & \textbf{0.77} & \textbf{0.701} & \textbf{0.851} & \textbf{0.672} \\
\bottomrule 
    \end{tabular}%
    }

\end{table*}%
\end{savenotes}


\subsection{Evaluation Results} 

\subsubsection{Inductive+Deductive case} First, we present the detailed results 
for the most challenging setting 
({\bf I+D}) for LUBM and NELL in \cref{tab:IDcaseLUBMNELL} and  Figure~\ref{fig:dataAug}.

Based on the average accuracy of the models across all query shapes reported in \cref{tab:IDcaseLUBMNELL}, the improvements of the proposed ontology-aware adaptations of \qbox and \cqd are evident. 
For LUBM \obox trained using the $\mi{onto}$ strategy improves the \qbox baseline by almost 50\%, while in case of \cqd, 54\% enhancement is achieved. 
For NELL similar behaviour is observed with the improvement of 
almost 20\% for \qbox, and 25\% for \cqd. Next, we discuss the impact of each of the proposed techniques for the E-OMQA task.

The first observation is that incorporating the ontology in the training data is crucial as both \qbox and \cqd trained in settings $\mi{gen}$ and $\mi{onto}$ 
yield significant improvements over the baselines.
Additional incorporation of specializations (setting $\mi{spec}$) does not seem 
to have a major impact though (see Figure~\ref{fig:dataAug}). On LUBM, for all models, 
the advantage of the ontology-driven query sampling (\ie, $\mi{onto}$ setting) 
is significant compared to $\mi{gen}$ setting. Remarkably, for LUBM $\cqd_\mi{onto}$, resp. $\cqdasr_\mi{onto}$ 
trained on less data than
 $\cqd_\mi{gen}$, resp. $\cqdasr_\mi{gen}$ results in higher accuracy. This shows that random query sampling is not adequate for E-OMQA.
 The ontology for NELL 
 is not expressive enough, thus, when generating training queries in the $\mi{onto}$ setting (see Table~\ref{tab:OntStatistics} for statistics)
we proceeded in a bottom-up fashion as follows: We randomly labeled query shapes which produce answers,
and constructed their generalizations as before; thus
 the settings $\mi{gen}$ and $\mi{onto}$ are similar,  
 but $\mi{onto}$ has 
 significantly less atomic queries, which explains why $\cqd_{\mi{gen}}$ outperforms $\cqd_{\mi{onto}}$ on NELL. 


On average ontology-aware models (i.e., \obox and \cqdasr) significantly outperform their baselines (i.e., \qbox and \cqd, resp.) for the majority of training data sampling strategies. This trend is more prominent for atom-based models on the complex LUBM ontology, and for query-based ones on NELL, which is less expressive. 

In Table~\ref{tab:baselines} we present the detailed results for the query rewriting over pre-trained embeddings. In order to evaluate  
this procedure, 
for each hard answer $a$  we take the 
best (i.e., minimum) ranking among all rankings generated by all queries in the rewriting of each test query. In other words, 
we take the minimal distance between the embedding of $a$ and all rewritings of $q$. Note that, for measuring the performance we use the pre-trained models $\mathit{Q2B_{plain}}$, and $\mathit{CQD_{plain}}$ obtained after 
450K training steps. 
Due to the reliance on particular query shapes of the respective models, the complete rewriting for each query is not guaranteed. In \cref{tab:baselines}, we present the results for this method compared to the $\mi{plain}$ setting.  Minor improvements of only at most 10\% are observed. 

\begin{figure*}[t]
\captionsetup[subfigure]{labelformat=empty}

    \resizebox{\textwidth}{!}{
\begin{subfigure}[t]{0.5\textwidth}
\pgfplotsset{ytick style={draw=none}, 
  legend columns=2, 
  legend style={
      draw=,/tikz/every even column/.append style={column sep=0.3cm},
      fill,
      at={(0.1,-0.1)},
      legend columns=2,
      legend cell align=left,
      anchor=north west
      },
  legend cell align={left}}
\begin{tikzpicture}
\begin{axis}[
    ybar,
    font=\LARGE,
    enlargelimits=0.15,
    legend style={at={(0.5,-0.15)},
      anchor=north,legend columns=-0.3, font=\Large},
    ylabel={Hits@3},
    symbolic x coords={plain,gen,spe,onto},
    xtick=data,
    ymajorgrids,
    legend style={nodes={scale=1.7, transform shape}}
    ]
\addplot[fill=gray] coordinates {(plain,0.699) 
(gen,0.661) (spe,0.651) (onto,0.583)};
\addplot[fill=red] coordinates {(plain,0.667) 
(gen,0.633) (spe,0.638) (onto,0.634)};
\addplot[fill=teal!40] coordinates {(plain,0.668) (gen,0.655) (spe,0.653) (onto,0.594)};
\addplot[fill=orange] coordinates {(plain,0.562) (gen,0.577) (spe,0.577) (onto,0.574)};
\legend{\qbox,\cqd,\obox,\cqdasr}
\end{axis}
\end{tikzpicture}
\caption{\huge LUBM: test case \tstA}
\label{lubm:tstA}
\end{subfigure} \hfil
\begin{subfigure}[t]{.5\textwidth} 
\begin{tikzpicture}
\begin{axis}[
    ybar,
    font=\LARGE,
     enlargelimits=0.15,
    legend style={at={(0.5,-0.15)},
      anchor=north,legend columns=-0.3, font=\Large},
    ylabel={Hits@3},
    symbolic x coords={plain,gen,spe,onto},
    xtick=data,
    ymajorgrids,
    legend style={nodes={scale=1.7, transform shape}}
    ]
\addplot[fill=gray] coordinates {(plain,0.253) 
(gen,0.506) (spe,0.506) (onto,0.818)};
\addplot[fill=red] coordinates {(plain,0.174) 
(gen,0.427) (spe,0.436) (onto,0.861)};
\addplot[fill=teal!40] coordinates {(plain,0.276) (gen,0.493) (spe,0.497) (onto,0.838)};
\addplot[fill=orange] coordinates {(plain,0.685) (gen,0.778) (spe,0.793) (onto,0.830)};
\legend{\qbox,\cqd,\obox,\cqdasr}
\end{axis}
\end{tikzpicture}
\caption{\huge LUBM: test case \tstB}
\label{lubm:tstB}
\end{subfigure} \hfil \quad
\begin{subfigure}[t]{0.5\textwidth}
\begin{tikzpicture}
\begin{axis}[
    ybar,
    font=\LARGE,
     enlargelimits=0.15,
    legend style={at={(0.5,-0.15)},
      anchor=north,legend columns=-0.3, font=\Large},
    ylabel={Hits@3},
    symbolic x coords={plain,gen,spe,onto},
    xtick=data,
    ymajorgrids,
    legend style={nodes={scale=1.7, transform shape}}
    ]
\addplot[fill=gray] coordinates {(plain,0.266) 
(gen,0.220) (spe,0.220) (onto,0.220)};

\addplot[fill=red] coordinates {(plain,0.326) (gen,0.278) (spe,0.279) (onto,0.286)};
\addplot[fill=teal!40] coordinates {(plain,0.244) 
(gen,0.228) (spe,0.228) (onto,0.226)};
\addplot[fill=orange] coordinates {(plain,0.255) 
(gen,0.259) (spe,0.259) (onto,0.271)};
\legend{\qbox,\cqd,\obox,\cqdasr}
\end{axis}
\end{tikzpicture}
\caption{\huge NELL: test case \tstA}
\label{nell:tstA}
\end{subfigure} \hfil
\begin{subfigure}[t]{.5\textwidth}
\begin{tikzpicture}
\begin{axis}[
    ybar,
    font=\LARGE,
     enlargelimits=0.15,
    legend style={at={(0.5,-0.15)},
      anchor=north,legend columns=-0.3, font=\Large},
    ylabel={Hits@3},
    symbolic x coords={plain,gen,spe,onto},
    xtick=data,
    ymajorgrids,
    legend style={nodes={scale=1.7, transform shape}}
    ]
\addplot[fill=gray] coordinates {(plain,0.521) 
(gen,0.734) (spe,0.734) (onto,0.725)};
\addplot[fill=red] coordinates {(plain,0.598) 
(gen,0.953) (spe,0.953) (onto,0.591)};
\addplot[fill=teal!40] coordinates {(plain,0.664)             (gen,0.744) (spe,0.745) (onto,0.748)};
\addplot[fill=orange] coordinates {(plain,0.708)             (gen,0.949) (spe,0.949) (onto,0.902)};
\legend{\qbox,\cqd,\obox,\cqdasr}
\end{axis}
\end{tikzpicture}
\caption{\huge NELL: test case \tstB}
\label{nell:tstB}
\end{subfigure} 
    }
    \caption{Comparison of $\qbox$,$\obox$, $\cqd$ and \cqdasr in each training setting for test cases (\textbf{I}) and (\textbf{D})}
    \label{fig:plotststIandD}
\end{figure*}

\subsubsection{Inductive case} Next, we present the average HITS@3 metric for the inductive {\bf I} test case (see ~\cref{fig:plotststIandD}).
For {\bf I} ontology-injection methods do not yield any improvement, which 
is expected, since ontologies cannot handle missing edges and facts in a KG that are not inferred from the data using ontological reasoning. 

\subsubsection{Deductive case} For the test case {\bf D}, when the ontology is simple (e.g., NELL), CQD and Q2B are able to more or less learn to apply the ontology rules when they are explicitly injected in the training set. Moreover, the results on NELL in the $\mi{plain}$ setting show that rule enforcement is also competitive for  deductive reasoning. For query-based models the best performance is achieved by combining ontology-driven sampling with rule enforcement, while for atom-based models, the inclusion of generalizations seems to be already sufficient.

For expressive ontologies, such as LUBM, the ontology-driven query sampling is crucial 
for optimal performance. For query-based models the best results are achieved when the ontology-driven query sampling is
combined with rule enforcement, while for atom-based models rule enforcement does not seem to be necessary.


\renewcommand{\arraystretch}{1.2}
\begin{table}[t]\centering \small
	\setlength{\tabcolsep}{10pt}
 	\caption{Avg. HITS@3 for QA of shapes 1p, 2p, 3p, 2i, 3i using rewriting on top of pre-trained $\mathit{{plain}}$ model vs the $\mathit{{plain}}$ model. 
	}
	\label{tab:baselines}
	\begin{tabular}{cc cc c} \toprule
		\multirow{2}{*}{ \textbf{Models}}  & \multicolumn{2}{c}{\bf Test Case \tstB} & \multicolumn{2}{c}{\bf Test Case \tstC} \\ \cmidrule(lr){2-3} \cmidrule(lr){4-5}
		& LUBM     & NELL  & LUBM   & NELL  \\ \midrule
       $\qbox_\mi{plain} $ & 0.189 & 0.617 & 0.193 & 0.539  \\
		\qboxrew   & 0.248 & 0.683 & 0.261 & 0.639 \\ \midrule \midrule
		\textbf{Gain} & +0.059 & +0.066 & +0.068 & + 0.1 \\ \midrule
		
		$\cqd_\mi{plain}$  & 0.225 & 0.656 & 0.231& 0.621 \\
		\cqdrew  & 0.228 & 0.743 & 0.249 & 0.708  \\ \midrule \midrule
		\textbf{Gain} & +0.003 & +0.087 & +0.018 & + 0.087 \\ \bottomrule
	\end{tabular} 
\end{table}

\section{Related Work}\label{sec:relwork}
The task of answering queries that involve multiple atoms using embedding techniques has recently received a lot of attention (see~\cite{DBLP:journals/corr/abs-2303-14617} for overview).
The existing proposals can be divided into \emph{query-based}~(e.g., \cite{iclr/RenHL20,DBLP:conf/nips/RenL20,DBLP:conf/kdd/LiuDJZT21,DBLP:conf/www/ChoudharyRKSR21,DBLP:conf/aaai/KotnisLN21,DBLP:conf/nips/SunAB0C20,DBLP:conf/icml/Zhu0Z022,DBLP:conf/nips/ZhangWCJW21}) and \emph{atom-based} (e.g., \cite{DBLP:journals/corr/abs-2011-03459,DBLP:journals/corr/abs-2301-12313}).

The works \cite{DBLP:conf/uai/FriedmanB20} and \cite{DBLP:conf/aaai/BorgwardtCL19} study the relation between the problem of conjunctive QA 
in the embedding space and over probabilistic databases. 
Our work is different from the above proposals in that along with the data we also rely on 
ontologies.

Integration of ontologies into KG embeddings has been recently actively investigated, for instance, in~\cite{DBLP:conf/semweb/KrompassBT15,DBLP:conf/uai/MinerviniDRR17,DBLP:conf/kdd/HaoCYSW19,DBLP:conf/semweb/JainTG021,DBLP:conf/emnlp/GuoWWWG16,DBLP:conf/nips/Kazemi018,DBLP:conf/aaai/FatemiR019,boxe,DBLP:conf/semweb/XiongPTNS22} (see also~\cite{DBLP:journals/corr/abs-2202-07412}), but these works typically focus on the task of link prediction rather than 
query answering. Recently, a type-aware model (called TEMP) for query answering over incomplete KGs has been proposed~\cite{DBLP:conf/ijcai/HuGXLLP22}. While TEMP allows for the exploitation of the type information, to the best of our knowledge it cannot handle more complex ontological axioms, which are the focus of our work.

The capability of embeddings to model hierarchical data has been explored in several works, e.g.,~\cite{DBLP:conf/akbc/PatelDB0VM20,kr/Gutierrez-Basulto18}.
Another relevant direction is concerned with the construction of models for $\mathcal{EL}$ ontologies in the embedding space~\cite{DBLP:journals/corr/abs-1902-10499}.
While 
the above works are related, 
they do not touch upon the problem of OMQA, studied in this work. 

The problem of ontology-mediated query answering 
has been considered in the area of knowledge representation and reasoning (see \eg \cite{DBLP:journals/ki/SchneiderS20a} for an overview), but available methods, e.g.\cite{DBLP:journals/corr/abs-1111-0049,aaai/EiterOSTX12}, only focus on 
logic-based deductive reasoning, but do not aim at predicting missing links in knowledge graphs using machine learning approaches. 

\section{Conclusion}\label{sec:conc}


We have presented methods for Embedding-based Ontology Mediated Query Answering (E-OMQA) that operate in the embedding space to enable simultaneous inductive and deductive reasoning over the incomplete data. 
Experiments show that embedding-based methods for query answering applied naively or combined with query rewriting techniques are not effective. At the same time, our ontology-aware extensions of  
 the popular 
 models for embedding-based 
 QA and the proposed ontology-driven training strategies yield promising results on the novel benchmarks that we introduce for the considered task.

 For future work we plan to study the effectiveness of our methods for embedding-based ontology mediated query answering for other more complex query forms~\cite{DBLP:conf/nips/RenL20,DBLP:conf/nips/WangYS21,DBLP:journals/corr/abs-2304-07063}, e.g., queries with negation, as well as evaluate the proposed approach for the cases when the ontology is more expressive. 


\section*{Acknowledgments}
Pasquale was partially funded by the European Union’s Horizon 2020 research and innovation programme under grant agreement no. 875160, ELIAI (The Edinburgh Laboratory for Integrated Artificial Intelligence) EPSRC (grant no. EP/W002876/1), an industry grant from Cisco, and a donation from Accenture LLP; and is grateful to NVIDIA GPU donations. This work was partially funded by the European project SMARTEDGE (grant number 101092908).

\bibliographystyle{ACM-Reference-Format}
\bibliography{ref}



\end{document}